\documentclass{article} 
\usepackage{colm2024_conference}

\usepackage{microtype}
\usepackage{hyperref}
\usepackage{url}
\usepackage{booktabs}

\usepackage{bbding}
\usepackage{amsmath}
\usepackage{amssymb}
\usepackage{arydshln}
\usepackage{bbm}
\usepackage{booktabs}
\usepackage{CJKutf8}
\usepackage{combelow}
\usepackage{enumitem}
\usepackage{epsfig}
\usepackage{fancybox}
\usepackage{float}
\usepackage{graphicx}
\usepackage{hyperref}
\usepackage{multirow}
\usepackage{multicol}
\usepackage{pgfplots}
\usepackage{pifont}
\usepackage{setspace}
\usepackage{subfigure}
\usepackage{tabularx}
\usepackage{url}
\usepackage{xspace}
\usepackage{wrapfig}

\newcommand{\our}{MetaIE\xspace}

\title{\our: Distilling a Meta Model from LLM for All Kinds of Information Extraction Tasks}

\author{Letian Peng, Zilong Wang, Feng Yao, Zihan Wang$^*$, Jingbo Shang\thanks{$\ $  Corresponding authors. } \\
University of California, San Diego \\
  \texttt{\{lepeng, zlwang, fengyao, ziw224, jshang\}@ucsd.edu}
  }

\colmfinalcopy 
\begin{document}

\maketitle

\begin{abstract}
    Information extraction (IE) is a fundamental area in natural language processing where prompting large language models (LLMs), even with in-context examples, cannot defeat small LMs tuned on very small IE datasets.
We observe that IE tasks, such as named entity recognition and relation extraction, all focus on extracting \emph{important information}, which can be formalized as a label-to-span matching.
In this paper, we propose a novel framework \our to build a small LM as meta-model by learning to extract ``important information'', i.e., the meta-understanding of IE, so that this meta-model can be adapted to all kind of IE tasks effectively and efficiently.
Specifically,  \our obtains the small LM via a symbolic distillation from an LLM following the label-to-span scheme.
We construct the distillation dataset via sampling sentences from language model pre-training datasets (e.g., OpenWebText in our implementation) and prompting an LLM to identify the typed spans of ``important information''.
We evaluate the meta-model under the few-shot adaptation setting. 
Extensive results on 13 datasets from 6 IE tasks
confirm that
\our can offer a better starting point for few-shot tuning on IE datasets
and outperform other meta-models from
    (1) vanilla language model pre-training,
    (2) multi-IE-task pre-training with human annotations, 
    and (3) single-IE-task symbolic distillation from LLM.
Moreover, we provide comprehensive analyses of \our, such as the size of the distillation dataset, the meta-model architecture, and the size of the meta-model.\footnote{Code, datasets, and model checkpoints: \url{https://github.com/KomeijiForce/MetaIE}.}

\end{abstract}

\section{Introduction}

Large language models (LLMs), such as ChatGPT~\citep{chatgpt}, benefit from vast amount of training data and have demonstrated exceptional performance across various areas through in-context learning (ICL)~\citep{icl-survey}. 
However, when it comes to information extraction (IE), LLMs, even with ICL examples, struggle to compete with smaller LMs (e.g., BERT~\citep{bert} and RoBERTa~\citep{roberta}) fine-tuned on very small training sets \citep{x-ner, llm-re, llm-ee}.
This is usually regarded as a limitation of LLMs in following a specific extraction scheme~\citep{llm-ie-survey}. 
Meanwhile, it is worth mentioning that conducting auto-regressive inference with LLMs is expensive and time-consuming, hindering their application in conducting IE over large corpora.

\begin{figure}[t]
    \centering
    \includegraphics[width=\linewidth]{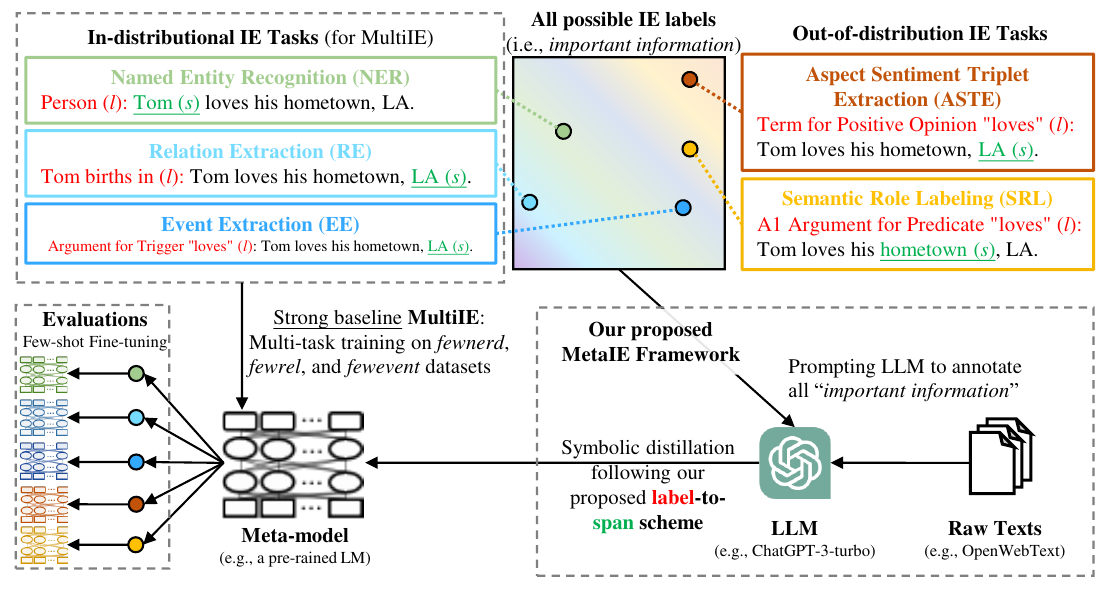}
    \vspace{-3mm}
    \caption{An overview of different transfer learning schemes involved in the experiments. }
    \label{fig:intro}
    \vspace{-5mm}
\end{figure}

We observe that IE tasks, such as named entity recognition (NER) and relation extraction (RE), all focus on extracting \emph{important information}, which can be formalized as \emph{label-to-span} instructions.
Specifically, all IE tasks can be decomposed as several instructions such as ``\emph{given an IE label ($l$), extract a span from the input text}'' (Figure~\ref{fig:intro}), where $l$ can be (1) \emph{Person}, \emph{Location}, \emph{Organization} in NER to recognize entities or (2) \emph{Tom births at} in RE to verify if there is a certain relation between two entities by checking the other entity can be recognized or not.
Following these label-to-span instructions, LLMs can handle all kinds of IE tasks and return imperfect yet semantically reasonable answers.
To this end, we argue that LLMs can be distilled into meta-models for IE which can quickly fine-tuned on few-shot training sets for better task-specific performance. 

In this paper, we propose a novel framework \our to build a small LM as a meta-model by learning to extract ``important information'', i.e., the meta-understanding of IE, and we show that this meta-model can be adapted to all kind of IE tasks effectively and efficiently.
Some prior work have built meta-models for a specific IE tasks, e.g., UniversalNER~\citep{universalner} explores the potential of building a meta-model for NER tasks. 
Our work is more ambitious at a larger scope for all IE tasks. 

\our obtains the small LM via a symbolic distillation~\citep{symbolic-distill} from an LLM following the label-to-span scheme.
We construct the distillation dataset via sampling sentences from language model pre-training datasets and prompting an LLM to identify the typed spans of ``important information''.
In particular, we implement this idea with $100,000$ sentences from the OpenWebText corpus~\citep{Gokaslan2019OpenWeb}, which contains various webpage texts and is also a subset of the popular language model pre-training dataset.
We feed these sentences to GPT-3.5-turbo for identifying ``important information'', which is then used to distill small LMs. 
It is worth mentioning that \our is applicable to all types of small LMs and one only needs to convert the label-span pairs following the corresponding labeling scheme (e.g., BIO sequence labeling for encoders like RoBERTa, seq2seq labeling for encoder-decoders like BART).

Our evaluation focuses on the few-shot learning ability of the meta-model for different IE tasks. 
We mainly compare \our with meta-models from (1) vanilla language model pre-training, (2) multi-IE-task pre-training with human annotations, and (3) single-IE-task symbolic distillation from LLM.
Large-scale datasets for NER, RE, and event extraction (EE) tasks are used in single-IE-task and multi-IE-task pre-training, therefore, these datasets shall be considered as \emph{in-task-distributional} for these two methods.
For a more comprehensive evaluation, we further include \emph{out-of-task-distributional datasets} from (1) semantic role labeling (SRL)~\citep{srl-task}, (2) aspect-based sentiment analysis (ABSA)~\citep{DBLP:conf/semeval/PontikiGPPAM14}, and (3) aspect-sentiment triplet extraction (ASTE)~\citep{aste-task}, totaling 13 datasets across 6 IE tasks.
In our experiments,
\our generally achieves the best performance, only \emph{very occasionally} losing to task-specific distillation on some in-task-distributional datasets.
This demonstrates that \our is a strong and efficient method to distill the meta-understanding of IE from LLMs into small LMs. 
Remarkably, distilling from the LLM-produced dataset following the traditional human annotation schemes performs poorly.
Therefore, the success of \our, rather than from purely using LLMs, shall also come from our label-to-span scheme. 

We have conducted comprehensive analyses of \our.
We study the scaling-up rules to investigate the model and dataset size boundaries in obtaining the meta-understanding of IE. 
We showcase the diversity of the types of important information in the \our distillation dataset. 
We show that the RoBERTa with sequence labeling framework is the best meta-model architecture compared with sequence-to-sequence and decoder-only models, at a similar scale. 

Our contributions are three-fold:
\begin{itemize}[nosep,leftmargin=*]
\item We are the first to build a small LM as a meta-model for all kinds of IE tasks.
\item We propose a novel label-to-span scheme that unifies all IE tasks and applies symbolic distillation to distill the meta-understanding from an LLM to a small LM.
\item We have a rigorous experiment design, which covers various IE tasks and meta-model methods. Comprehensive experiment results support the intuitive expectation and advantage of our \our.
\end{itemize}

\section{Related Works}

\subsection{Information Extraction}

Information extraction (IE) is one of the most popular and vital domains in natural language processing. Early IE systems are generally developed for a single IE dataset like NER \citep{DBLP:conf/aclnews/SantosG15}, RE \citep{DBLP:conf/acl/KatiyarC16}, or EE \citep{DBLP:conf/acl/ChenXLZ015}. Due to the gap between the label sets and annotation styles of different IE datasets, few-shot IE frameworks \citep{fewnerd,fewrel,fewevent} are proposed to quickly learn models on new datasets. The IE models are pre-trained on a large scale of IE labels and then transferred to the target domain by fine-tuning on few examples. With the emergence of LLMs, researchers have started to train LMs on multiple IE tasks with unified formats \citep{UIE,tanl}. LLMs fine-tuned for general purpose \citep{chatgpt,llama-2} have also shown strong potential to understand new IE tasks with their instruction-following ability. However, these LLMs still lag behind supervised models \citep{llm-ie-survey}, potentially due to the difficulty of specifying the required pattern for extraction in different datasets. Moreover, the cost of LLMs limits their application to IE on a large corpus. This paper aims to transfer the meta-understanding of IE from LLMs to lighter-weight models, which produce a flexible model with high adaptability to any target IE task.

\subsection{Model Distillation}

Model distillation \citep{model-distillation,distill-survey} is the process of transferring knowledge from large models (teacher models) to small ones (student models). Traditional distillation optimizes the similarity between logits produced by the teacher and student models \citep{model-distillation,qakd,tqkd}. Symbolic distillation \citep{symbolic-distill,symbolic-cot-distill,nova-comet} for language models learns a student model on texts generated by the teacher model. In comparison with traditional distillation, symbolic distillation allows the student model to focus on one aspect of the teacher model \citep{symbolic-distill}, which can be some high-level ability, such as chain-of-thought reasoning \citep{symbolic-cot-distill}, with much smaller model size. For IE, symbolic model distillation has been successfully applied for an IE subtask, NER \citep{universalner}, which distills an NER model that can extract entities in a broad domain. This paper aims to distill the cross-IE task ability of LLMs, i.e., meta-understanding of IE and proposes a meta-model that can effectively learn IE tasks with few examples. 

\subsection{Meta Learning}

Meta-learning \citep{meta-learning} enables the models to learn new tasks better, i.e., stronger transfer learning ability. MAML \citep{maml} proposes a framework to learn a better starting point for few-shot learning by utilizing multiple datasets for loss updating. Reptile \citep{reptile}, similar to MAML, simplifies the meta-learning algorithm by performing stochastic gradient descent not only within each task but also across tasks, making it more efficient and easier to implement. The Prototypical Networks method \citep{prototype} employs a distance-based classification approach, where it learns a metric space in which classification can be performed by computing distances to prototype representations of each class. While most meta-learning methods are experimented on classification tasks, pre-training on multiple datasets \citep{fewnerd} and prototypical networks \citep{protoner} have been applied for IE. While these methods focus on specific IE tasks like NER, we aim to optimize a starting point for general IE tasks by distilling from LLMs.

\section{Our \our Framework}

\subsection{Label-to-span Scheme}

We formalize the IE task as given an IE label $l$ (e.g., \emph{Person} in NER), extracting a span $s$ from a sentence $X = [x_1, \cdots, x_n]$. 
The span $s$ can be represented as $x_{i:j}$ including the words from $i$-th to $j$-th. 
Denoting the IE process as a mapping $f_{IE}(\cdot)$, it can be represented as $s=f_{IE}(X|l)$. Machine learning-based methods aim to learn the mapping by optimizing a model $M_\theta$ with parameter $\theta$. 
For a specific IE task (e.g., NER), the IE label set $\mathcal{L}^{(Task)}$ will contain $l$ falling inside the task label, i.e., $(l\in \mathcal{L}^{(Task)})$.
Based on the general definition of IE, the general IE label set $\mathcal{L}^{(IE)}$ can be any textual description, thus $\forall \text{Task}, \mathcal{L}^{(Task)}\subset \mathcal{L}^{(IE)}$. 

In this paper, we aim to learn a meta-model that can be easily adapted to different IE tasks. 
In the current practice of IE, the ``meta-model'' is generally pre-trained in a single IE task with a large number of labels ($\mathcal{L}^{(pt)} \subset \mathcal{L}^{(Task)}$). 
Then, the meta-model can be fine-tuned on few-shot examples to quickly adapt to different downstream IE datasets \emph{in the same task}, such that $\mathcal{L}^{(ft)} \subset \mathcal{L}^{(Task)}$. 
We expand this learning scheme to a general meta-model that works for all existing and potentially new IE tasks.
To achieve this goal, our intuition is to pre-train the model to learn the label-to-span mapping with the label set approximating the general IE label distribution $\mathcal{L}^{(pt)} \sim \mathcal{L}^{(IE)}$. 
As the label sets of all IE tasks are subsets of $\mathcal{L}^{(IE)}$, our meta-model will enjoy an efficient transfer to all IE tasks.

\subsection{Distillation Dataset Construction}

\begin{wrapfigure}{r}{0.45\textwidth}
  \vspace{-12mm}
  \begin{center}
    \includegraphics[width=0.44\textwidth]{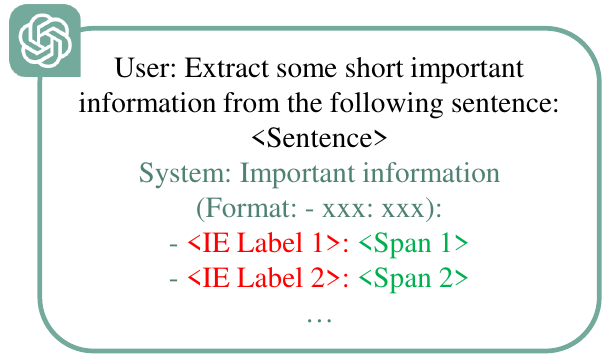}
  \end{center}
  \vspace{-3mm}
  \caption{The prompt used in our experiments to build the dataset for symbolic distillation.}
  \label{fig:prompt}
  \vspace{-3mm}
\end{wrapfigure}

To apply a symbolic distillation of the meta-understanding of IE from LLMs, we prompt LLMs to create data for distillation by querying them to extract ``important information'' from texts as shown in Figure~\ref{fig:prompt}. 
Our expectation for the dataset is to cover as many $l$ as possible to approximate the broad $\mathcal{L}^{(IE)}$ set to better distill the meta-model for all kinds of IE tasks.
We query LLMs to annotate some raw corpora $\mathcal{X}$ to build the MetaIE dataset. 
Given each $X\in \mathcal{X}$, the LLM is instructed to generate a series of $(l, s)$ pairs. 
We do not set any limitation to $l$ to better approximate the broad $\mathcal{L}^{(IE)}$ set. 

\paragraph{Implementation} 
We select the paragraphs from OpenWebText~\citep{Gokaslan2019OpenWeb}, Since OpenWebText it is a popular dataset used in language model pre-training, we are not introducing new texts. 
We split the paragraphs by sentences and only use the first sentence of each paragraph for a higher diversity and to avoid the ambiguity caused by coreference.
The LLM is instructed to formalize all $(l, s)$ pairs in the prompting output as ``- \underline{Place} $(l)$: \underline{New York} $(s)$'', which are extracted by regular expression matching. 
Considering there might be multiple spans returned for $l$, we split the span by conjunctions like comma. 

\begin{table*}
\centering
\small
\scalebox{1.0}{
\begin{tabular}{p{1.5cm}p{11.5cm}}
\toprule
$n$-gram (Count) & Example IE Labels (Relative Frequency) \\
\midrule
$1$-gram (270k) &  Location (7.73\%), Event (4.67\%), Action (4.24\%), Topic (3.57\%), Subject (3.25\%), Person (2.71\%), Date (2.70\%), Source (2.44\%)\\
\midrule
$2$-gram (44.5k) &  Target audience (1.27\%), Time period (0.998\%), Individuals involved (0.992\%), Action taken (0.877\%), Political affiliation (0.762\%), Parties involved (0.758\%), Release date (0.697\%), TV show (0.686\%)\\
\midrule
$3$-gram (16.9k) & Source of information (2.02\%), Cause of death (1.17\%), Call to action (1.02\%), Date of birth (0.739\%), Date and time (0.727\%), Date of death (0.562\%), Type of content (0.337\%), Reason for arrest (0.325\%)\\
\midrule
$4$-gram (7.39k) &  Purpose of the bill (0.325\%), Location of the incident (0.271\%), Name of the person (0.271\%), Number of people killed (0.203\%), Number of people affected (0.189\%), Content of the bill (0.162\%), Number of people arrested (0.149\%), Source of the information (0.149\%)\\
\midrule
$\geq 5$-gram (5.37k) &  Dates of birth and death (0.13\%), Age at the time of death (0.112\%), Total number of votes cast (0.0931\%), Feature: Auschwitz through the Lens of the SS (0.0931\%), Number of people on board (0.0745\%), Name of the person involved (0.0745\%), Date and time of publication (0.0745\%), Action taken by President Obama (0.0745\%)\\
\bottomrule
\end{tabular}
}
\caption{Example IE Labels, Counts, and Relative Frequency in our constructed symbolic distillation dataset, grouped by the number of tokens.} 
\label{tab:query_stats}
\end{table*}

Table~\ref{tab:query_stats} shows some statistics and example results of the labels returned by the LLM, illustrating a broad spectrum of IE domains, ranging from simple entities and events to complex relationships and contexts. 
The diversity in the $n$-gram categories
showcases the model's ability to capture a wide array of query types.
This variety underscores the comprehensive coverage and nuanced understanding that LLMs bring to the task of generating queries across different facets of the IE domain.

\subsection{Distillation Framework}

We illustrate the distillation with a sequence labeling model~\citep{DBLP:conf/aclnews/SantosG15} that suits well for encoder-based language models (e.g., RoBERTa \citep{roberta}).
Given a sequence of words $X = [x_1, \cdots, x_n]$, the sequence labeling model will tag each word by outputting $Y = [y_1, \cdots, y_n]$. 
Following the traditional BIO labeling scheme, $y_i$ will be $B$ (begin), $I$ (inner), and $O$ (none). The model is trained on word tagging and the tags are decoded into spans by searching sequences that begin with $B$ and continue by $I$. In traditional sequence labeling models, the $B$ and $I$ tags generally consist of label information such as $B$-place or $I$-person. 
In our case, we formalize the tagging in a query-dependent way since the model needs to handle arbitrary queries. 
We attach the label information as a prefix like ``place: '' to the beginning of the input text. 
The input text is then labeled by the BIO scheme, where the span label is indicated in the prefix. 
Finally, the BIO sequences are used to fine-tune the sequence labeling models.
This distillation process can also be adapted to Seq2Seq encoder-decoder models and Causal LM-based decoder-only models.
We use sequence labeling models for the main experiment based on their empirical advantage in IE tasks, which we also empirically find support in the analysis in Section~\ref{sec:framework}.

\section{Experiments}

\subsection{IE Tasks and Datasets}

To deeply delve into the differences between different model distillation or meta-learning methods, we include a wide variety of tasks: 
\begin{enumerate}[nosep,leftmargin=*]
    \item Named Entity Recognition (\textbf{NER}) extracts named entities with their labels from texts.
        We include $6$ NER datasets that was studied in~\cite{tner}, i.e., (1) \textbf{CoNLL2003}, (2) \textbf{BioNLP2004}, (3)   \textbf{WNUT2017}, (4) \textbf{MIT-Movie}, (5) \textbf{MIT-Restaurant}, (6) \textbf{BC5CDR}, which covers various domains: news, medical, social media, and reviews.
    \item Relation Extraction (\textbf{RE}) extracts named entities, and in addition, identifies the relationships between them. 
        We include $2$ popular datasets, (1) \textbf{ADE}~\citep{ADE} and (2) \textbf{CoNLL2004}~\citep{conll2004} representing RE on medical and news domain. 
        We evaluate the performance of RE models on both relation detection and the detection of entities involved in the relations.
    \item Event Extraction (\textbf{EE}) extracts event triggers and their arguments. We use the standard \textbf{ACE2005} dataset~\citep{ace2005multilingual} for EE evaluation. 
        We compare the model performance on both event trigger detection (T) evaluation task and trigger-augment pair detection (A) evaluation task. 
    \item Semantic Role Labeling (\textbf{SRL}) extracts predicates (verbs) and their arguments.
        We select the \textbf{CoNLL2005}~\citep{srl-task} dataset for SRL. 
        We follow previous works to learn backbone LMs on samples from the Brown training dataset and then test them on Brown and WSJ test datasets.
    \item Aspect-based Sentiment Analysis (\textbf{ABSA}) extracts aspect terms and the sentiment polarity towards them. 
        We select \textbf{SemEval2014}~\citep{DBLP:conf/semeval/PontikiGPPAM14} as the dataset for ABSA, with its two subsets: \textbf{14res} and \textbf{14lap} including reviews about restaurants and laptops. 
    \item Aspect Sentiment Triplet Extraction (\textbf{ASTE}) extracts aspect terms and the corresponding opinion terms that contain the sentiment polarity towards them.
        We use the same \textbf{SemEval2014} dataset as for ABSA, on which aspect-sentiment triplets are further annotated by~\citet{aste-task}.
\end{enumerate}

For a fair comparison, we formalize all those tasks as $s=f_{IE}(X|l)$, which can be found in the Appendix~\ref{apdx:l2s}.
For each task, we query each possible label to extract $(l, s)$ pairs.
For spans conflicting with each other, as we run label-wise extractions, we only keep the one with a higher BI sequence probability. 
For tasks that extractions are dependent on each other (e.g., RE, EE, SRL, ASTE), we follow~\cite{tanl} to run multi-stage extractions for these tasks. 
As ACE2005 involves too many labels, we report the unlabeled performance on detecting the triggers and arguments for all methods for comparison.

\subsection{Evaluation Metric: Few-shot Fine-tuning Performance}

We use the few-shot fine-tuning performance on all IE tasks to evaluate the meta-model's quality. 
Specifically, all methods in our evaluation will provide us a backbone LM. 
We then conduct few-shot fine-tuning from the training dataset for fine-tuning with sample details in Appendix~\ref{apdx:few_detail}.
Finally, we evaluate them on the test dataset using the micro F1 score as the evaluation metric. 
For multi-task pre-training baselines, tasks without large-scale annotations (SRL, ABSA, ASTE) are \textbf{out-of-distribution tasks}.

The default backbone LM we used for fine-tuning is \texttt{RoBERTa-Large}~\citep{roberta}, which is a traditional bidirectional encoder used for learning IE tasks formalized as sequence tagging. 
The learning rate is set to $2\times 10^{-5}$ with AdamW~\citep{AdamW} as the optimizer and a cosine annealing learning rate scheduler~\citep{SGDR}. 
We fine-tune the backbone LM with batch size $64$ for a single epoch to avoid overfitting.

\subsection{Compared Methods}

We first include a comparison with the teacher model \textbf{GPT-3.5-turbo} via \textbf{LLM Prompting} with in-context learning (\textbf{ICL}). 
For ICL, we provide $5$ examples in the prompt of our query.
Based on previous discoveries on LLM-based IE \citep{x-ner, llm-re, llm-ee}, we shall expect that fine-tuned small LMs work better than the LLM.

We compare our \textbf{\our} with a variety of methods from the following three categories
\begin{enumerate}[nosep,leftmargin=*]
    \item \textbf{Vanilla} LM fine-tuning (\textbf{FT}), i.e., directly using the vanilla pre-trained LM as the backbone LM in fine-tuning.
    \item \textbf{Task-level} Meta-learning (\textbf{ML})\textbf{+FT}. It is expected to have a strong performance to other datasets in the same IE task but poor generalization to other IE tasks.
        \begin{itemize}[nosep,leftmargin=*]
            \item \textbf{Transfer (Human)} is a baseline that trains the backbone LM on large-scale human annotations of a specific IE task. Specifically, we use \emph{FewNerd}~\citep{fewnerd} for NER, \emph{FewRels}~\citep{fewrel} for RE, and \emph{FewEvents}~\citep{fewevent} for EE. 
            \item \textbf{Transfer (LLM)} uses the same datasets in \textbf{Transfer (Human)} but queries the LLM to annotate them following the human workflow. This baseline aims to compare the quality of annotation from humans and LLMs following the conventional annotation schema. 
            \item \textbf{Task Distillation} distills from LLMs by querying answers for specific IE tasks. We implement this by providing in-context task-specific examples to control the LLM-produced data similar to the label IE task. The input texts are set to be the same as MetaIE to avoid bias.
            \item \textbf{NER Distillation} applies the model distilled following \textbf{Task Distillation} but tests them on non-NER tasks to evaluate its cross-task transferability.
        \end{itemize}
    \item \textbf{IE-level} Meta-learning (\textbf{ML})+\textbf{FT} aims to learn an IE model with strong transferability to all IE tasks. Our \textbf{\our} also falls into this category.
        \begin{itemize}[nosep,leftmargin=*]
            \item \textbf{MultiIE} merges the multiple human-annotated IE datasets (\emph{FewNerd}, \emph{FewRels}, \emph{FewEvents}) to train a backbone LM, which represents a multi-task baseline with human annotations. 
            \item \textbf{MAML}~\citep{maml} is a traditional meta-learning baseline that merges gradients on different datasets to build a model that can be quickly transferred to these datasets. We use the datasets in \textbf{MultiIE} for \textbf{MAML} in the experiment.
        \end{itemize}
\end{enumerate}

For all baselines, the data number for meta-learning is controlled to the same as \our by sampling towards a fair comparison. 

\subsection{Result}

\begin{table*}
\small
\centering
\scalebox{.72}{
\begin{tabular}{l|l|cccccc}
\toprule
\multirow{2}*{Category}& \multirow{2}*{Method} & \multicolumn{6}{c}{NER}\\
 &  & ConLL$2003$ & BioNLP$2004$ & WNUT$2017$ & MIT-Movie & MIT-Restaurant & BC5CDR \\
\midrule
{LLM Prompting} & ICL & $59.68$ & $48.08$ & $36.51$ & $46.08$ & $60.62$ & $59.82$ \\
\midrule
\multirow{1}*{FT} & Vanilla & $32.58$ & $36.06$ & $33.87$ & $57.65$ & $63.40$ & $18.15$ \\
\midrule
\multirow{4}*{Task-level ML+FT} & Transfer  \\
& $\quad$ Human & $71.61$ & $54.58$ & $43.15$ & $\textbf{64.80}$ & $69.17$ & $72.02$ \\
& $\quad$ LLM & $67.74$ & $45.62$ & $45.36$ & $59.59$ & $69.19$ & $73.14$ \\
& Task Distillation & $\textbf{74.86}$ & $\textbf{56.18}$ & $\textbf{50.09}$ & $\textbf{65.70}$ & $\textbf{71.48}$ & $71.01$ \\
\midrule
\multirow{3}*{IE-level ML+FT}& MultiIE & $63.94$ & $52.47$ & $44.29$ & $58.43$ & $69.38$ & $71.20$ \\
& MAML & $66.97$ & $53.09$ & $46.14$ & $60.57$ & $68.86$ & $72.58$ \\
& MetaIE & $71.49$ & $\textbf{55.76}$ & $44.33$ & $\textbf{65.64}$ & $\textbf{71.33}$ & $\textbf{75.21}$ \\
\midrule
\midrule
\multirow{2}*{Category} & \multirow{2}*{Method} & \multicolumn{2}{c}{RE (NER)} & \multicolumn{2}{c}{RE} &  \multicolumn{2}{c}{EE}\\
& & ADE & CoNLL$2004$& ADE & CoNLL$2004$ & ACE$2005$ (T) & ACE$2005$ (A)  \\
\midrule
{LLM Prompting} & ICL & $63.55$ & $58.47$ & $39.02$ & $31.34$ & $60.47$ & $28.79$ \\
\midrule
\multirow{1}*{FT} & Vanilla & $25.97$ & $62.13$ & $15.67$ & $33.52$ & $67.46$ & $32.86$ \\
\midrule
\multirow{5}*{Task-level ML+FT} & Transfer \\
& $\quad$ Human & $41.56$ & $\textbf{69.27}$ & $20.53$ & $37.51$ & $\textbf{72.79}$ & $35.77$ \\
& $\quad$ LLM & $35.43$ & $66.93$ & $14.35$ & $35.07$ & $65.17$ & $34.86$ \\
& Task Distillation & $66.99$ & $68.66$ & $\textbf{41.92}$ & $41.58$ & $67.34$ & $34.56$ \\
& NER Distillation & $67.35$ & $\textbf{69.88}$ & $32.73$ & $35.68$ & $66.17$ & $32.86$ \\
\midrule
\multirow{3}*{IE-level ML+FT} & MultiIE & $53.26$ & $\textbf{69.14}$ & $18.23$ & $39.65$ & $71.16$ & $35.23$ \\
& MAML & $56.95$ & $\textbf{69.28}$ & $38.65$ & $42.07$ & $68.22$ & $35.84$ \\
& MetaIE & $\textbf{69.29}$ & $\textbf{69.47}$ & $\textbf{40.43}$ & $\textbf{43.50}$ & $69.85$ & $\textbf{36.83}$ \\
\midrule
\midrule
\multirow{2}*{Category}& \multirow{2}*{Method} & \multicolumn{2}{c}{SRL} & \multicolumn{2}{c}{ABSA} & \multicolumn{2}{c}{ASTE} \\
& & Brown & WSJ & 14RES & 14LAP & 14RES & 14LAP  \\
\midrule
{LLM Prompting} & ICL & $28.79$ & $31.56$ & $53.04$ & $35.62$ & $58.94$ & $44.87$ \\
\midrule
\multirow{1}*{FT} & Vanilla & $52.59$ & $56.47$ & $24.46$ & $10.32$ & $39.17$ & $41.50$ \\
\midrule
\multirow{1}*{Task-level ML+FT} & NER Distillation & $43.65$ & $51.29$ & $10.77$ & $11.21$ & $40.06$ & $38.40$ \\
\midrule
\multirow{3}*{IE-level ML+FT} & MultiIE & $52.26$ & $56.63$ & $38.22$ & $35.28$ & $24.91$ & $40.49$ \\
& MAML & $52.69$ & $56.23$ & $40.22$ & $34.45$ & $30.83$ & $40.95$ \\
& MetaIE & $\textbf{54.50}$ & $\textbf{58.49}$ & $\textbf{50.96}$ & $\textbf{39.71}$ & $\textbf{43.30}$ & $\textbf{43.10}$ \\
\bottomrule
\end{tabular}
}
\caption{Few-shot transferring performance (F1 score) of different meta-learning sources on IE tasks. \textbf{Bold:} Performance of the \emph{small LM} that is not significantly different from the best one. $(p < 0.05)$} 
\label{tab:main}
\end{table*}

The result from our experiments is presented in Table~\ref{tab:main}. The vanilla model is poorly transferred by fine-tuning to all kinds of IE tasks. The model with meta-learning on a single IE task, NER, is only well-transferred to other NER datasets but poorly-transferred to other IE tasks. Among IE-level meta-learning methods, the MultiIE model can be transferred to in-domain IE tasks with outstanding performance but still fails to be transferred to out-of-domain IE tasks, either with regular pre-training or meta-learning frameworks like MAML. In contrast to all these baselines, our MetaIE shows a strong transferability to all IE tasks, especially on out-of-domain tasks for MultiIE. Thus, the experiment results are highly consistent with our claim in IE task transferability that wider pre-training label set $\mathcal{L}^{(IE)}$ will enable macro transferability of the model to all IE tasks. 

Besides the main discovery, we can also observe that LLM-based meta-learning outperforms the pre-training on human annotation. Take NER as an instance, while both label sets satisfy $\mathcal{L} \subset \mathcal{L}^{(NER)}$, the $\mathcal{L}$ proposed by LLMs is much more diverse than the fixed set in human annotated datasets, which again verifies the importance of the label distribution, even in task-specific distillation.

The comparison with the teacher model also shows the student model generally outperforming the teacher model under few-shot supervision. Thus, we conclude fine-tuning a distilled student IE model to perform better than inference by the teacher LLMs with few-shot in-context examples. This further verifies the advantage of model distillation in meta-learning which enables more efficient and effective transfer.

\section{Further Analysis}

\subsection{Size Analysis}

We explore how the scale of the student model or the data number affects the distillation quality. For the model scale, we compare among \texttt{RoBERTa-Small}, \texttt{RoBERTa-Base}, and \texttt{RoBERTa-Large}. For the data scale, we increase the sampling size to $640K$ and pre-train the student model with different amounts of data.

\begin{figure}
    \centering
    \includegraphics[width=\linewidth]{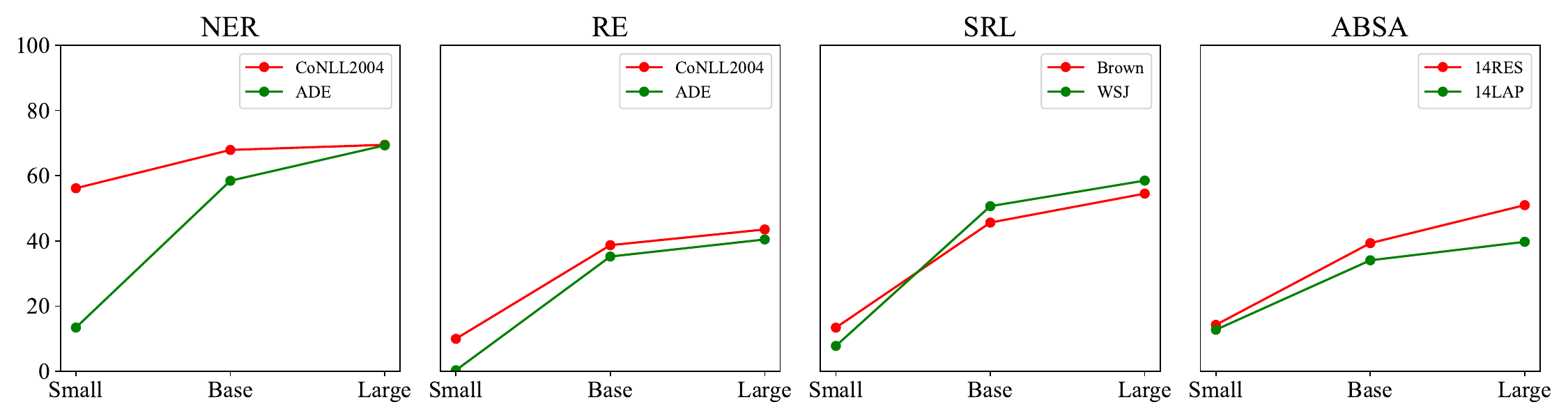}
    \caption{The size analysis of the student model scale on different IE tasks and domains.}
    \label{fig:scale_size}
    \vspace{-5mm}
\end{figure}

\begin{figure}
    \centering
    \scalebox{0.8}{\includegraphics[width=\linewidth]{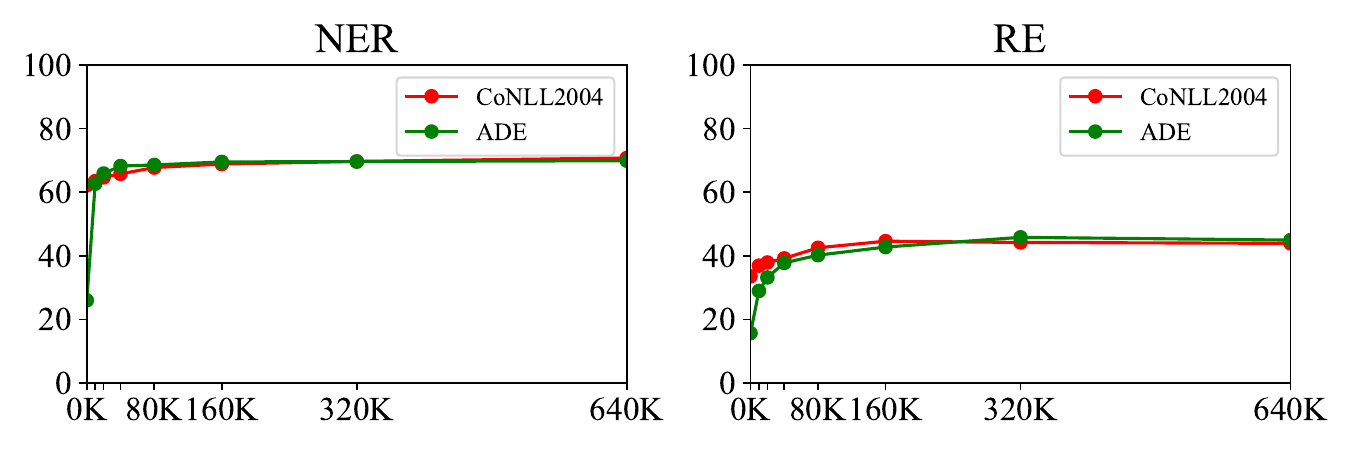}}
    \caption{The size analysis of the distillation data scale on different IE tasks and domains.}
    \label{fig:scale_data}
    \vspace{-5mm}
\end{figure}

The analysis of \textbf{model size} is presented in Figure~\ref{fig:scale_size}, we can observe the performance of a student model can be scaled up by more parameters. Also, for simple tasks (like NER) with a general domain (like CoNLL2004), a tiny student model is competent for the distillation. However, for specific domains or complex tasks, the student model needs more parameters for generalization. 

The analysis of \textbf{data size} is presented in Figure~\ref{fig:scale_data}, we observe the existence of a threshold between $80K \sim 160K$ to endow the student model with the meta-understanding of IE. Also, a small amount of meta data like $10K$ can significantly benefit the transferring.

\subsection{Distillation Framework Comparison}
\label{sec:framework}

We compare student models following different distillation frameworks (because of their architectures) to investigate how this factor affects the distillation effectiveness. 

\textbf{Seq2Seq} implements the distillation by learning to extract a group of spans based on the IE label as in the distillation dataset. We include two Seq2Seq models: \texttt{BART-Large} \citep{bart} and \texttt{T5-Base} \citep{t5}, which contain the same scale of parameters as in the \texttt{RoBERTa-Large} in our previous experiments. 

\textbf{CausalLM} is similar to \textbf{Seq2Seq} but only uses the decoder model instead of the encoder-decoder as in \textbf{Seq2Seq}. We also include two CausalLM-based models with similar parameter scales: \texttt{GPT2-Medium} \citep{gpt2} and \texttt{OPT-350M} \citep{opt}.

We also include another sequence labeling model \texttt{BERT-Large-Cased} \citep{bert} as a baseline to explore the influence of the backbone model quality on the learning performance. For all models, we pre-train them using our MetaIE dataset with the same hyperparameters. 

\begin{table*}
\small
\centering
\scalebox{.83}{
\begin{tabular}{l|l|cccccc}
\toprule
Framework & Model & ConLL$2003$ & BioNLP$2004$ & WNUT$2017$ & MIT-Movie & MIT-Restaurant & BC5CDR \\
\midrule
\multirow{2}*{Seq-Labeling} & BERT & $63.01$ & $52.39$ & $32.71$ & $61.75$ & $62.50$ & $66.24$ \\
& RoBERTa & $\textbf{71.49}$ & $\textbf{54.88}$ & $44.33$ & $\textbf{65.64}$ & $\textbf{71.33}$ & $\textbf{75.21}$ \\
\midrule
\multirow{2}*{Seq2Seq} & BART & $\textbf{71.39}$ & $47.18$ & $\textbf{46.74}$ & $62.76$ & $67.98$ & $65.90$ \\
& T5 & $64.01$ & $42.35$ & $40.74$ & $55.05$ & $53.60$ & $38.67$ \\
\midrule
\multirow{2}*{CausalLM} & GPT & $57.20$ & $37.29$ & $36.89$ & $52.14$ & $60.46$ & $61.03$ \\
& OPT & $52.39$ & $37.64$ & $34.48$ & $53.07$ & $53.59$ & $52.86$ \\
\bottomrule
\end{tabular}
}
\caption{Comparison between different frameworks on MetaIE distillation.} 
\label{tab:framework}
\vspace{-5mm}
\end{table*}

We compare the performance of different distillation frameworks on NER as an example and the result is demonstrated in Table~\ref{tab:framework}. Sequence labeling models perform the best in few-shot transfer learning, which indicates their advantage in the distillation of meta-understanding of IE. This can be attributed to the consistency of sequence labeling with the extraction nature. We thus conclude distilling IE knowledge to a traditional sequence labeling model is better than those popular generative models. Between sequence labeling models, RoBERTa outperforms BERT, showing a better student model also benefits the distillation procedure. 

\section{Limitation Discussion}

\textbf{Efficiency} The efficiency of the unified label-to-span will be $O(|\mathcal{L}^{(Task)}|)$, which is lower than the traditional $O(1)$ (number of LM forwarding) BIO sequence labeler with label information in the labeling result. This will limit the application of our model to cases where $|\mathcal{L}^{(Task)}|$ is large. This efficiency is a trade-off for the ability to process any IE label, which enables the fast transfer of the BIO model to different IE tasks. 

\textbf{Bias in LLM-proposed labels} As pointed out in previous works \citep{llm-bias, bias-ai-content}, LLMs have biases in their responses. This can also be observed in the statistics of our distillation dataset. Thus, the small meta-model might also inherit the bias and have better transferability to labels that LLMs prefer than others.

\section{Conclusions and Future Work}

This paper presents a novel approach for distilling the meta-understanding of IE from LLMs into more efficient, smaller language models through a synthesized dataset, MetaIE. Our findings indicate that this method not only enhances the adaptability and efficiency of smaller models but also outperforms existing single-task and multi-task distillation methods in various IE tasks. The success of MetaIE underscores the potential of leveraging LLM's meta-understanding to improve the performance and versatility of smaller models in complex tasks, offering a promising direction for future research in model distillation and IE. Future work will explore a better way for meta-learning by distilling from LLMs and other meta-tasks can be trained based on distillation. 

\bibliography{colm2024_conference}
\bibliographystyle{colm2024_conference}

\newpage

\appendix
\section{Label-to-Span Formalization}
\label{apdx:l2s}

\textbf{NER}

\textbf{Person:} \underline{John/B Smith/I} loves/O his/O hometown/O ,/O Los/O Angeles/O

\textbf{RE}

\textbf{Person:} \underline{John/B Smith/I} loves/O his/O hometown/O ,/O Los/O Angeles/O

\textbf{John Smith births in:} John/O Smith/O loves/O his/O hometown/O ,/O \underline{Los/B Angeles/I}

\textbf{EE}

\textbf{Trigger:} John/O Smith/O \underline{loves/B} his/O hometown/O ,/O Los/O Angeles/O

\textbf{Argument for Trigger ``loves'':} John/O Smith/O loves/O his/O hometown/O ,/O \underline{Los/B Angeles/I}

\textbf{SRL}

\textbf{Verb:} John/O Smith/O \underline{loves/B} his/O hometown/O ,/O Los/O Angeles/O

\textbf{A1 Argument for Verb ``loves'':} John/O Smith/O loves/O his/O \underline{hometown/B} ,/O Los/O Angeles/O

\textbf{ABSA}

\textbf{Positive Term:} John/O Smith/O loves/O his/O hometown/O ,/O \underline{Los/B Angeles/I}

\textbf{ASTE}

\textbf{Positive Opinion:} John/O Smith/O \underline{loves/B} his/O hometown/O ,/O Los/O Angeles/O

\textbf{Aspect for Opinion ``loves'':} John/O Smith/O loves/O his/O hometown/O ,/O \underline{Los/B Angeles/I}

\section{Few-shot Details}
\label{apdx:few_detail}
\paragraph{NER} samples $5$-shot examples that contain a certain type of entity for each entity type.
\paragraph{RE} samples $5$-shot examples that contain a certain type of relation for each relation type.
\paragraph{EE} samples $5\%$ examples from the original training dataset.
\paragraph{SRL} samples $50$-shot examples from the original training dataset.
\paragraph{ABSA} samples $5$-shot examples that contain terms with a certain sentiment polarity for each sentiment polarity type.
\paragraph{ASTE} samples $5$-shot examples that contain aspect-opinion triplet with a certain sentiment polarity for each sentiment polarity type.

\end{document}